\documentclass{article}

\usepackage[dblblindworkshop, nonatbib, final]{neurips_2025}

\usepackage[utf8]{inputenc} 
\usepackage[T1]{fontenc}    
\usepackage{hyperref}       
\usepackage{url}            
\usepackage{booktabs}       
\usepackage{amsfonts}       
\usepackage{nicefrac}       
\usepackage{microtype}      
\usepackage{xcolor}         
\usepackage{graphicx}
\usepackage{tabularx}
\usepackage{wrapfig}
\usepackage{amsmath}
\usepackage{float}
\usepackage{listings}
\usepackage{authblk}
\usepackage{setspace}

\usepackage{biblatex}
\title{Evaluating Long-Context Reasoning in LLM-Based WebAgents}
\workshoptitle{LAW 2025: Bridging Language, Agent, and World Models }

\author[1]{Andy Chung}
\author[1]{Yichi Zhang}
\author[2]{Kaixiang Lin}
\author[2]{Aditya Rawal}
\author[2]{Qiaozi Gao}
\author[1]{Joyce Chai}

\affil[1]{University of Michigan}
\affil[2]{Amazon}

\begin{document}

\maketitle

\begin{abstract}
As large language model (LLM)-based agents become increasingly integrated into daily digital interactions, their ability to reason across long interaction histories becomes crucial for providing personalized and contextually aware assistance. However, the performance of these agents in long context scenarios, particularly for action-taking WebAgents operating in realistic web environments, remains largely unexplored. This paper introduces a benchmark for evaluating long context reasoning capabilities of WebAgents through sequentially dependent subtasks that require retrieval and application of information from extended interaction histories.
We develop a novel evaluation framework that simulates multi-session user interactions by injecting irrelevant task trajectories between dependent subtasks, creating contexts ranging from 25,000 to 150,000 tokens. Through extensive evaluation of four popular models, Claude-3.7, GPT-4.1, Llama 4, and o4-mini, we observe a dramatic performance degradation as context length increases, with success rates dropping from 40-50\% in baseline conditions to less than 10\% in long context scenarios. Our detailed error analysis reveals that agents primarily fail due to getting stuck in loops and losing track of original task objectives. We further propose an implicit RAG approach that provides modest improvements by generating task-relevant summaries, though fundamental limitations in long context reasoning persist. These findings highlight critical challenges for deploying WebAgents in realistic, long-term user interaction scenarios and provide insights for developing more robust agent architectures capable of maintaining coherent task execution across extended contexts.
\end{abstract}

\section{Introduction}
Action-taking large language model (LLM)-based agents are rapidly becoming ubiquitous in our daily lives. These assistants extend their capabilities beyond the chatbot environment, making tangible impacts through their ability to operate in the digital realm. For instance, an action-taking agent can go beyond merely suggesting vacation plans--it can go online to book your flights and reserve your hotel as well.

As these assistants become more capable, users will increasingly expect them to recognize and adapt to their preferences and intentions. A user may ask the assistant to find the highest-rated movie on IMDb and add it to your watchlist. Days later, they might simply say, “Buy that movie we saved earlier,” and expect it to understand the reference. 

While there have been many studies on the personalization of LLMs for chatbots and information retrieval assistants \cite{lyu2024llmrecpersonalizedrecommendationprompting, Zhang2024PersonalizedLR, zhang2025personalizationlargelanguagemodels}, studies exploring the personalization capabilities of LLM-based action-taking agents are lacking.

We study WebAgents in the live open Internet, one of the most information-rich environments on earth. Its usage is ubiquitous in North America with 96\% of the population online. This environment also enables an unparalleled variety of tasks such as shopping, social media, banking, entertainment, or research. This diversity enables us to design the experiment to closely reflect real-world human behavior. Despite the challenges and limitations of studying live environments, this approach enables us to observe how these agents perform under realistic conditions.

We explore the capabilities of these types of agents in a realistic long context situation by performing the following:
\begin{itemize}
    \item We introduce a new benchmark for evaluating web agents on long-horizon, sequential tasks. The benchmark includes a dataset of sequentially dependent subtasks and a novel methodology that simulates long-term interaction by injecting noisy historical trajectories of varying lengths into the agent's context.
    \item We conduct a comprehensive evaluation of state-of-the-art web agent models on our benchmark, showing significant performance degradation as context length increases. Through a detailed analysis of reasoning traces, we identify and categorize key failure modes related to the agent's planning and reasoning abilities in long-context settings.
    \item We propose an implicit Retrieval-Augmented Generation (iRAG) approach that improves task success rates by retrieving relevant information from the lengthy noisy context history.
\end{itemize}

Our results and error analysis show several key insights. Task success rates drop dramatically when noise is introduced, dropping by about 30\% from the case when there is no noise. Analysis of the reasoning traces show that the WebAgent tends to get stuck in loops, repeating the same ineffective action over and over again until the step limit is reached. The WebAgent's task efficiency drops sharply as context length increases with most cases reaching the step limit. Breaking down the instructions into sub-instructions can assist the WebAgent in completing tasks when the context is long.

\section{Related Work}
The meteoric rise of powerful large language models (LLMs) has shown strong capabilities in reasoning in dynamic open world environments such as the world wide web \cite{furuta_multimodal_2024,gur_real-world_2024,he_webvoyager_2024,yao2023reactsynergizingreasoningacting,zheng_gpt-4vision_2024}. These generally capable agents are able to complete complex tasks by breaking them down into multiple steps and then completing those steps in a sequential stepwise fashion.

Prefeval is a benchmark that evaluates a LLM's capability to adhere to user preferences in a long context conversational setting. The experimental setup involves long context retrieval of information in order to produce correct outputs. However, the setting mimics a QA type of instruction rather than a trajectory of an action-taking agent \cite{zhao2025llmsrecognizepreferencesevaluating}.

RealWebAssist is a benchmark that evaluates long-horizon web-based tasks under realistic scenarios and includes ambiguous instructions an user may give. The study also investigates the performance of a WebAgent at varying context lengths. However, the study only reports an analysis of performance at relatively short context lengths of up to 20 steps. A granular analysis of long-context interactions which could include hundreds of steps was not provided nor is there a detailed analysis of errors \cite{Ye2025RealWebAssistAB}.

\section{Long-Context Reasoning Benchmark}
\subsection{Problem Formulation}
\begin{wrapfigure}{r}{0.5\textwidth} 
    \centering
    \vspace{-10pt} 
    \includegraphics[width=0.48\textwidth]{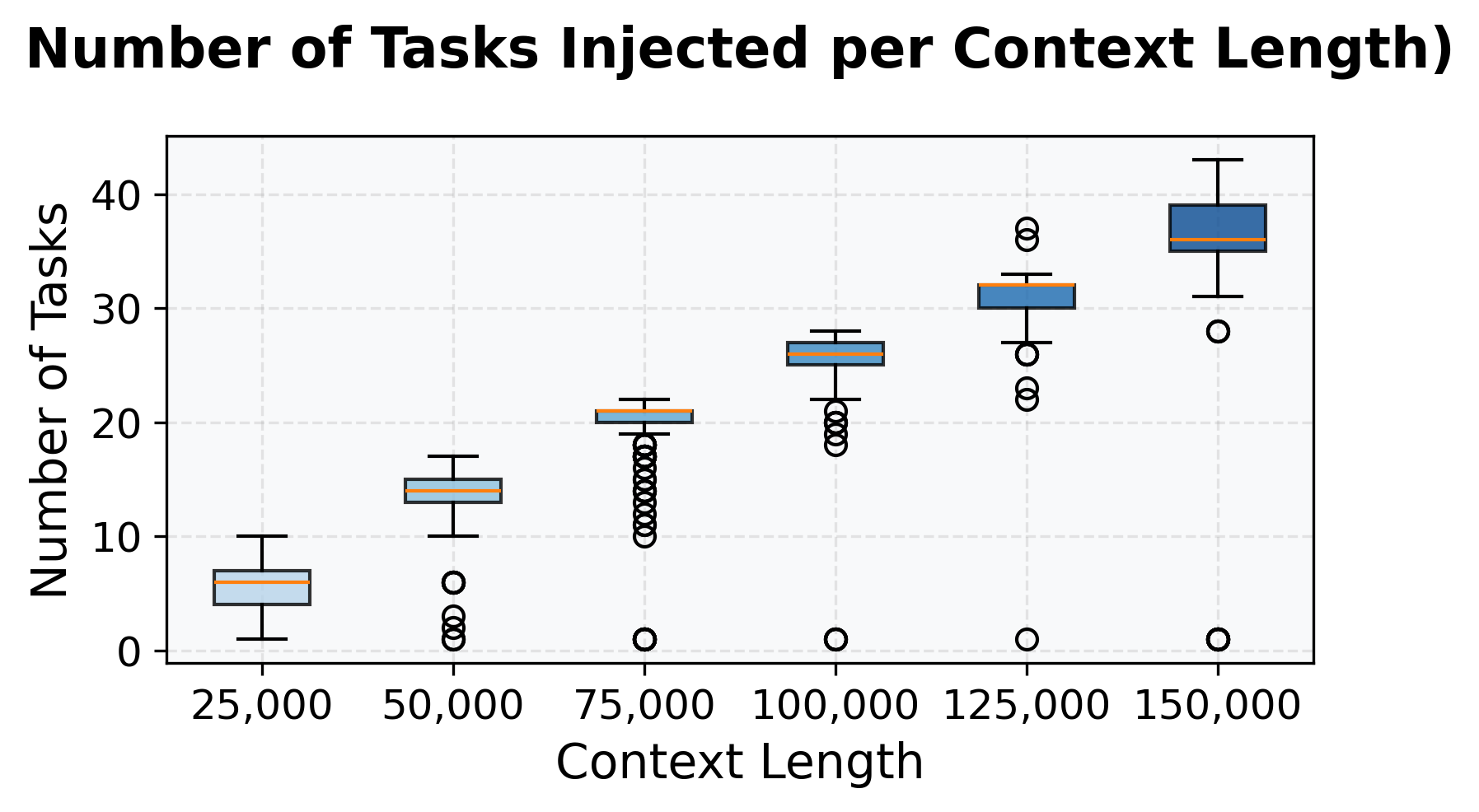}
    \caption{Number of tasks injected at each context length}
    \label{tasklengthfig}
    \vspace{-10pt} 
\end{wrapfigure}

Over time, our interaction history becomes a rich source of contextual information, enabling these agents to better interpret our intentions and personalize their responses. As our engagement with them deepens, both the variety and complexity of tasks naturally increase. To remain useful in our daily lives over the long term, an assistant must be able to reason across this expanding context.
For our benchmark, we setup a scenario to simulate a user's interaction over various periods of time. First we have the WebAgent perform subtask $A_1$. After the completion of subtask $A_1$, We inject irrelevant trajectories up to a certain absolute context length: [25,000, 50,000, 75,000, 100,000, 125,000, 150,000].  Each step in each trajectory contains a thought, action, reflection, and the current URL at that step. This simulates the passage of time as if users completed a number of tasks after completing $A_1$. The number of tasks injected depends on the target context length of the experiment, as shown in Figure \ref{tasklengthfig}. We then set the current URL of the agent to the last step of the last injected task to simulate the completion of that step. The agent is then asked to perform another subtask, $A_2$.  We also have a 'no noise' condition where no irrelevant tasks are injected as a baseline. \ref{tasklengthfig} shows the number of tasks injected at each context length. We use these results as our base case.

\subsection{Task Definition}
\begin{figure*}[t]
\begin{center}
\includegraphics[width=1\textwidth]{images/experimental_setup.png}
\end{center}
\label{experimental_setup}
\caption{Benchmark framework and evaluation overview. The left column shows the setup of the multi-session context. The middle column provides examples of the split subtasks as well as the (irrelevant) multi-session tasks. The right flowchart demonstrates benchmark and evaluation process for the trajectory the agent takes. }
\end{figure*}

Context consists of instruction, observation, thought, action, and reflection stored in memory as the interaction history. Over time, this interaction history becomes complex and long as more and more tasks are fulfilled by the agent. In order to study long context understanding over this interaction history, we divide a given task into two sequentially dependent subtasks. We formulate the task as follows:
\begin{itemize}
    \item[--] Let $A$ = Task
    \item[--] Let $A_1=$ sequential subtask of $A$
    \item[--] Let $A_2=$ second sequential subtask of $A$
\end{itemize}
$A_2$ must satisfy two properties:
\begin{itemize}
    \item[1.] Under-specification: The second sub-task must be under-specified by some attribute. For example, if $A_1$ was to find hotels near the airport, then $A_2$ could be to find restaurants near that airport. In this case, the attribute hotel from the first sub-task is not specified. This forces the LLM to recall that piece of information from a previously executed trajectory
    \item[2.] Unambiguous: $A_2$ must depend solely on $A_1$ in the sequence. As previously noted, additional tasks will be introduced into the context. These injected tasks must be orthogonal from $A_1$. If any injected task bears resemblance to $A_1$, the agent might justifiably base $A_2$ on the injected task rather than following $A_1$'s trajectory. To accurately assess the dependency chain, we must ensure that $A_1$ remains clearly distinct from the injected subtasks. Therefore, we omit any trajectory other than $A_1$ that may reasonably be used to complete $A_2$.

\end{itemize}

\subsection{Subtask Generation}
Initially, we attempted to generate a dataset that satisfies the properties above using a LLM. However, we found that while many of the tasks generated seem plausible, they end up being unexecutable and/or difficult to evaluate. This is likely due to the superficial understanding current SOTA foundational models have about website affordances. For example, one of the tasks generated was "goto expedia.com and navigate to the "Flights" Search Page ($A_1$). Then click the "Search" button without entering any information ($A_2$)." The task is plausible. However, if you were to actually try this, you would get an error saying the "leaving from" and "going to" fields are empty, rendering the task unexecutable. While there are ways to autonomously generate feasible tasks \cite{zhou2024proposeragentevaluatorpaeautonomousskilldiscovery}, we have the additional constraints of generating two sequentially dependent subtasks that need to be executable along with the properties mentioned above. Thus, we utilize an existing dataset from WebCanvas \cite{pan2024webcanvasbenchmarkingwebagents} to source our executable tasks. WebCanvas offers a collection of 543 annotated tasks (438 in the dev set, 105 in test set) spanning diverse domains, including travel, information retrieval, services, shopping, and entertainment.  The tasks are also curated to remove any time-sensitive metrics such as current temperature or tasks involving making a booking at a certain date. There is an average of 8 steps per tasks with a total of 4550 total steps across all tasks.

Although WebCanvas provides a large number of tasks, many of the tasks are no longer executable due to the dynamic nature of the live Internet. Therefore, to ensure the tasks are still executable, we curate a set of successful tasks by running the WebAgent across the entire dataset using Claude-3.5 and 3.7 as base models, yielding a total of 56 tasks. From the successful tasks, we then prompt a LLM to split the task into $A_1$ and $A_2$. We manually inspect $A_1$ to make sure it contains all the information needed to complete that part of the task and inspect $A_2$ to make sure they satisfy the under-specification requirement. To ensure that $A_2$ is unambiguous, we use a LLM to classify any task (other than $A_1$) that can be used to reasonably complete $A_2$.

\section{Evaluation}
We evaluate Claude-3.7, GPT-4.1, Llama-4-17B Maverick, and o4-mini on our benchmark due to their relatively strong performance on agentic tasks as well as their ability to support long context lengths. For Claude-3.7, we do not use thinking tokens. For all models except o4-mini, we use temperature = 0 to provide more stable results. Error bands are calculated using the standard error of proportions for binary variables. For continuous variables we use standard error of the mean.

\subsection{Task Success Criteria}
In our experiments, we found that evaluations that were purely judged by a LLM tend to fluctuate. This is also evidenced by previous studies such as \cite{he_webvoyager_2024}, and \cite{pan2024autonomousevaluationrefinementdigital}. The LLM may judge an incomplete task or a similar but incorrect final state as successful. Even with temperature set to 0, agents do not take the same trajectory given a task, which will affect the LLM's judgment. In practice, LLMs are not deterministic \cite{atil2025nondeterminismdeterministicllmsettings}, producing variations at the string level even with temperature = 0. This string level variances are exacerbated due the the long reasoning chains needed to complete multi-step agentic tasks. This variance is also noted in Anthropic's documentation \cite{anthropic_api_docs}. Additionally, the live Internet is dynamic in nature. For example, a different ad may be displayed on every visit to the website, affecting the state and context which in turn would affect the decisions of the agent and therefore the trajectory of a given task. Thus, we utilize key steps as described in WebCanvas for evaluation of the tasks rather than LLM based evaluations. Key steps are checkpoints that must be passed in the trajectory for a given task regardless of any variation in other steps of the trajectory. For example, the task "Add all dlc to cart for the game DOTA on steam" would have the following keysteps:  
\begin{itemize}
    \item[1.] Goto steampowered.com
    \item[2.] Load the landing page for the DOTA 2 game.
    \item[3.] Click on the "add all dlc to cart" button.
\end{itemize}
They are rule based, thus removing one layer of variation in our evaluations. For a task to be considered successful, all key steps for a given task must be successful. On average, there are 4.5 key steps per task.

\subsection{Implicit RAG (iRAG) Approach}
\begin{wrapfigure}{r}{0.5\textwidth} 
    \centering
    \vspace{-10pt} 
    \includegraphics[width=0.48\textwidth]{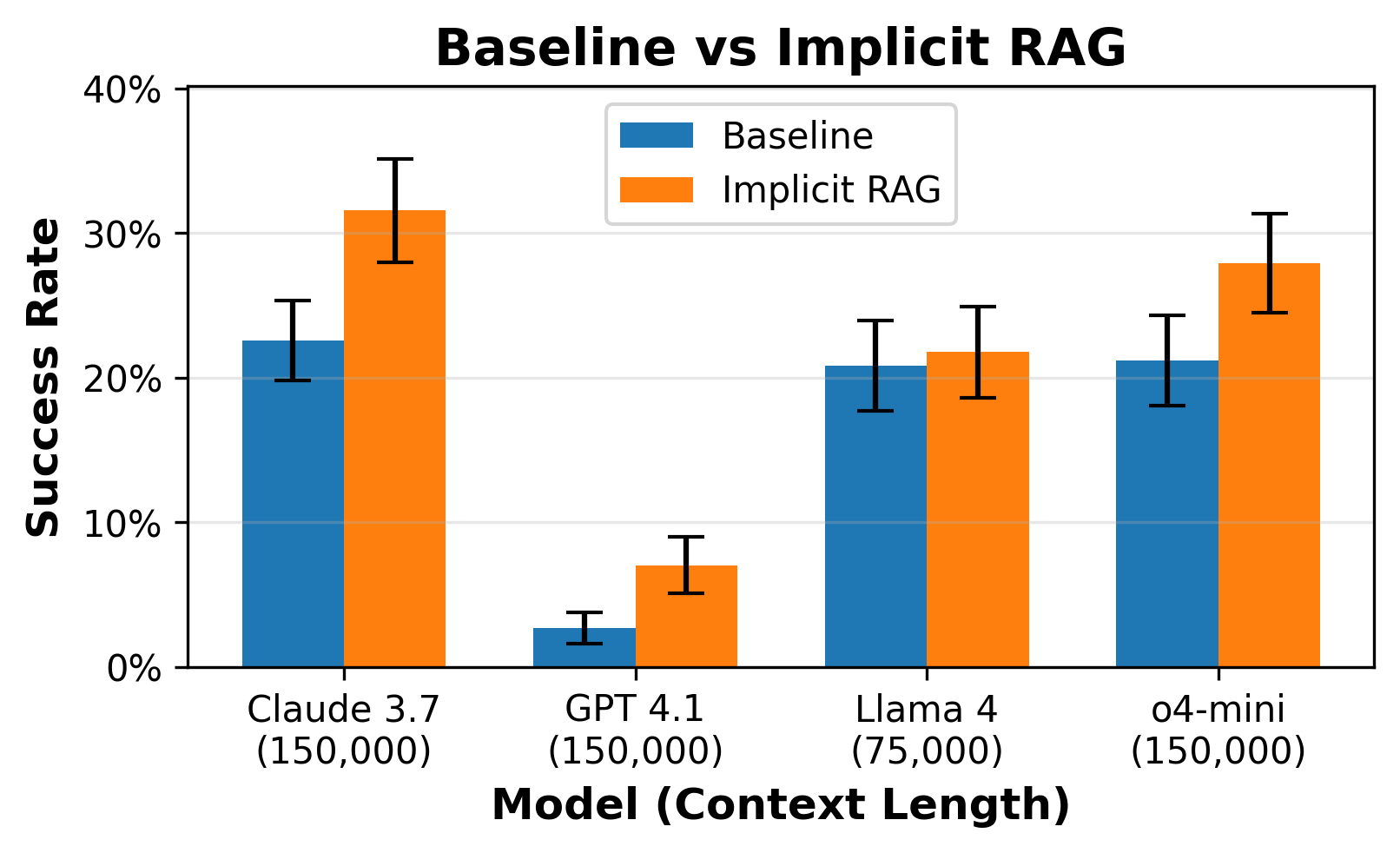}
    \caption{Comparison of performance of baseline vs implicit RAG at 150k context length.}
    \label{results}
    \vspace{-10pt} 
\end{wrapfigure}

Each step, the WebAgent is instructed to follow many instructions in one step including observation of the accessibility tree and the current screenshot, reasoning over the interaction history, utilizing feedback from the previous step, generating a thought with respect to the current state and the given task, generating the next action, identifying the correct web element to perform the action on, and responding in the specified structured output format. We hypothesize that the long context interferes with the agent's ability to follow these instructions, and crucially, retrieve relevant information needed to successfully complete a given task. Thus, we add a separate 'implicit RAG' step to assist the WebAgent. In this step, we simplify the instruction and ask the WebAgent to only generate a summary with respect to its current sub-task, state, and the provided interaction history. The intuition behind this approach is to break a complex instruction into sub-instructions, just like how breaking up a complex task into sub-tasks can make task completion easier. The summary is then appended to the context and the step is proceeded as before.

\section{Experiments}
\begin{wrapfigure}{r}{0.5\textwidth} 
    \centering
    \vspace{-10pt} 
    \includegraphics[width=0.48\textwidth]{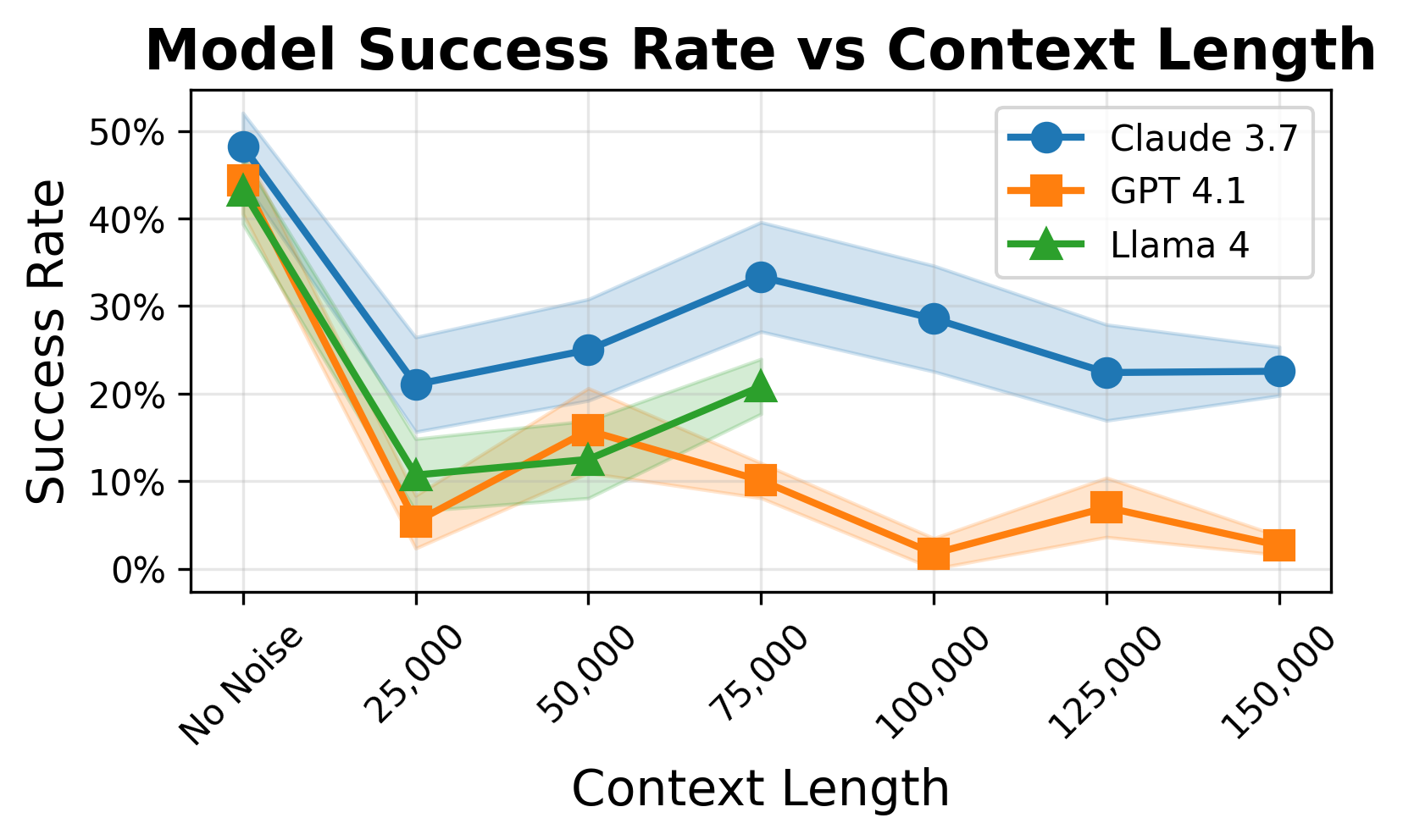}
    \caption{Success rate of task completion at varying context lengths}
    \label{results}
    \vspace{-10pt} 
\end{wrapfigure}

Our results, shown in Figure \ref{results}, demonstrate a sharp drop in performance once noise is injected into the trajectory dropping from 40-50\% to as low as $<10$\% success rate. This indicates that once retrieval is required of the WebAgent, the success rate drops dramatically. O4\- mini performs relatively well, indicating that reasoning models outperform non-reasoning models in long context reasoning tasks.

In the implicit RAG approach, we see an improvement in the performance of the WebAgent compared to the baseline for Claude-3.7, GPT-4.1, and o4-mini at 150k context length. This suggests that decomposing complex instructions into smaller sub-instructions enhances WebAgent's performance.

\begin{table}[h]
\centering
\begin{tabularx}{0.9\columnwidth}{|p{2cm}|X|X|}
  \hline
  \textbf{Context Length} & \textbf{o4-mini} & \textbf{GPT-4.1} \\
  \hline
  No Noise & \textbf{46.2\% ± 3.8\%} & 44.4\% ± 3.8\% \\
  150000   & \textbf{21.2\% ± 3.1\%} & 2.7\% ± 1.1\% \\
  \hline
\end{tabularx}
\caption{Performance comparison of o4-mini and GPT-4.1}
\end{table}

\begin{table}[htbp]
\centering
\begin{tabular}{llc}
\toprule
Model & Category & Percentage $\pm$ Std Error \\
\midrule
Claude 3.7 & False End & 16.37\% $\pm$ 1.47\% \\
 & Inefficient Progress & \textbf{34.97\% $\pm$ 2.65\%} \\
 & Loops & 16.67\% $\pm$ 1.66\% \\
 & Other Errors & 1.19\% $\pm$ 0.40\% \\
 & Success & 30.80\% $\pm$ 3.48\% \\
 \midrule
GPT 4.1 & False End & 6.90\% $\pm$ 1.09\% \\
 & Inefficient Progress & 32.74\% $\pm$ 5.63\% \\
 & Loops & \textbf{44.29\% $\pm$ 8.03\%} \\
 & Other Errors & 1.67\% $\pm$ 0.51\% \\
 & Success & 14.40\% $\pm$ 4.22\% \\
 \midrule
Llama 4 & False End & 11.61\% $\pm$ 3.24\% \\
 & Inefficient Progress & \textbf{39.96\% $\pm$ 3.80\%} \\
 & Loops & 21.21\% $\pm$ 1.37\% \\
 & Other Errors & 0.22\% $\pm$ 0.22\% \\
 & Success & 27.01\% $\pm$ 5.10\% \\
 \midrule
o4-mini & False End & 2.68\% $\pm$ 0.61\% \\
 & Inefficient Progress & 19.35\% $\pm$ 5.16\% \\
 & Loops & \textbf{43.15\% $\pm$ 2.86\%} \\
 & Other Errors & 0.89\% $\pm$ 0.40\% \\
 & Success & 33.93\% $\pm$ 6.17\% \\
\bottomrule
\end{tabular}
\caption{Error categories of tested models. Most common error modes are bolded.}
\label{tab:model_performance_pct}
\end{table}

In order to better understand why the WebAgents fail, we categorize the errors into the following categories:
\begin{itemize}
    \item False End: The WebAgent decides to end the task but has not completed all key steps.
    \item Step Limit
        \begin{itemize}
          \item Loops: The WebAgent reaches the step limit and is stuck in a loop.
          \item Inefficient Progress: The WebAgent is progressing in the task but has reached the step limit.
        \end{itemize}
    \item Other Errors: These include technical errors like timeouts, anti-bot measures, or LLM request rejections.
\end{itemize}

The breakdown of error categories does not show a clear trend with respect to context length. The majority of errors are step limit errors due to either being stuck in a loop or progressing the task inefficiently. Notably, GPT-4.1 tends to get stuck in a loop more often than Claude-3.7. This may indicate Claude models have additional mechanisms implemented to deal with situations where the input repeatability remains the same in a multi-turn setting. False ends happen less frequently than step limit errors.  Other errors are rare and include websites timing out for extended periods of time, getting blocked by anti-bot measures, or the LLM refusing the request due to the request being perceived as harmful.

Analysis of the reasoning traces indicates that once the WebAgent is stuck in a loop, it has a difficult time breaking out of it even when explicitly instructed to recognize the loop and attempt alternative actions. In most cases, it retries the same ineffective action over and over until the step limit is reached. However, there are some cases where the agent tries the same action a few times and then decides to try a different action, breaking it out of the loop.

Analysis of reasoning traces when the step limit is reached and it is not a loop shows, in many cases, the agent exploring the wrong parts of the website or on an entirely different domain than what it should be on. In this scenario, the agent is unable to realize it has deviated from the correct trajectory and searches various parts of the website or tries to go to a different website until it reaches the step limit.

Analysis of false ends show that in many cases, the agent ends up on a different website that accomplishes a similar goal despite the task asking for completion of the task on the given website. For example, the task may ask to find NBA stats on a player on sports.yahoo.com but the WebAgent ends up finding NBA stats on nba.com. In another instance, only partial information about the original subtask was retrieved. The WebAgent decides to end the task due to satisfaction of this partially retrieved information. For example, one of the tasks was to find cruises on www.carnival.com with Miami, FL as the port and the Bahamas as the destination. It was able to recall that it was supposed to search for cruises on www.carnival.com and returned to the correct domain, but did not recall that it was supposed to find the cruises for that specific route. The WebAgent decided to end the task after performing a generic cruise search. The WebAgent may also decide to end the task because it falsely concludes that it is impossible. This is due to it deviating from the correct trajectory. In the most egregious case, the agent completely loses track of the original task and thinks it has successfully completed the last injected task.

\subsection{Retrieval Performance Analysis}

\begin{table}[htbp]
\centering
\begin{tabular}{lcccc}
\toprule
\textbf{Model} & \textbf{TS, RF (\%)} & \textbf{TS, RS (\%)} & \textbf{TF, RS (\%)} & \textbf{TF, RF (\%)} \\
\midrule
Claude 3.7     & 0.0 ± 0.0  & 25.1 ± 2.0  & 26.9 ± 1.5  & \textbf{48.0 ± 2.3} \\
GPT 4.1        & 0.0 ± 0.0  & 7.0 ± 1.4   & 21.0 ± 4.4  & \textbf{72.0 ± 5.3} \\
Llama 4        & 0.0 ± 0.0  & 17.1 ± 2.8  & \textbf{42.9 ± 1.9}  & 40.0 ± 4.1 \\
o4-mini        & 0.0 ± 0.0  & 21.7 ± 1.2  & 37.4 ± 2.3  & \textbf{40.9 ± 3.2} \\
\bottomrule
\end{tabular}
\caption{Retrieval performance on the benchmark, aggregated across context lengths. Error modes are grouped by frequency, with the most prevalent types emphasized in bold. For Llama, evaluations were conducted up to a maximum context length of 75,000 tokens. Abbreviations: TS — Task Success; RS — Retrieval Success; TF — Task Failure; RF — Retrieval Failure.}
\label{tab:synthetic_error_catagories}
\end{table}

\begin{wrapfigure}{r}{0.5\textwidth} 
    \centering
    \vspace{-10pt} 
    \includegraphics[width=0.48\textwidth]{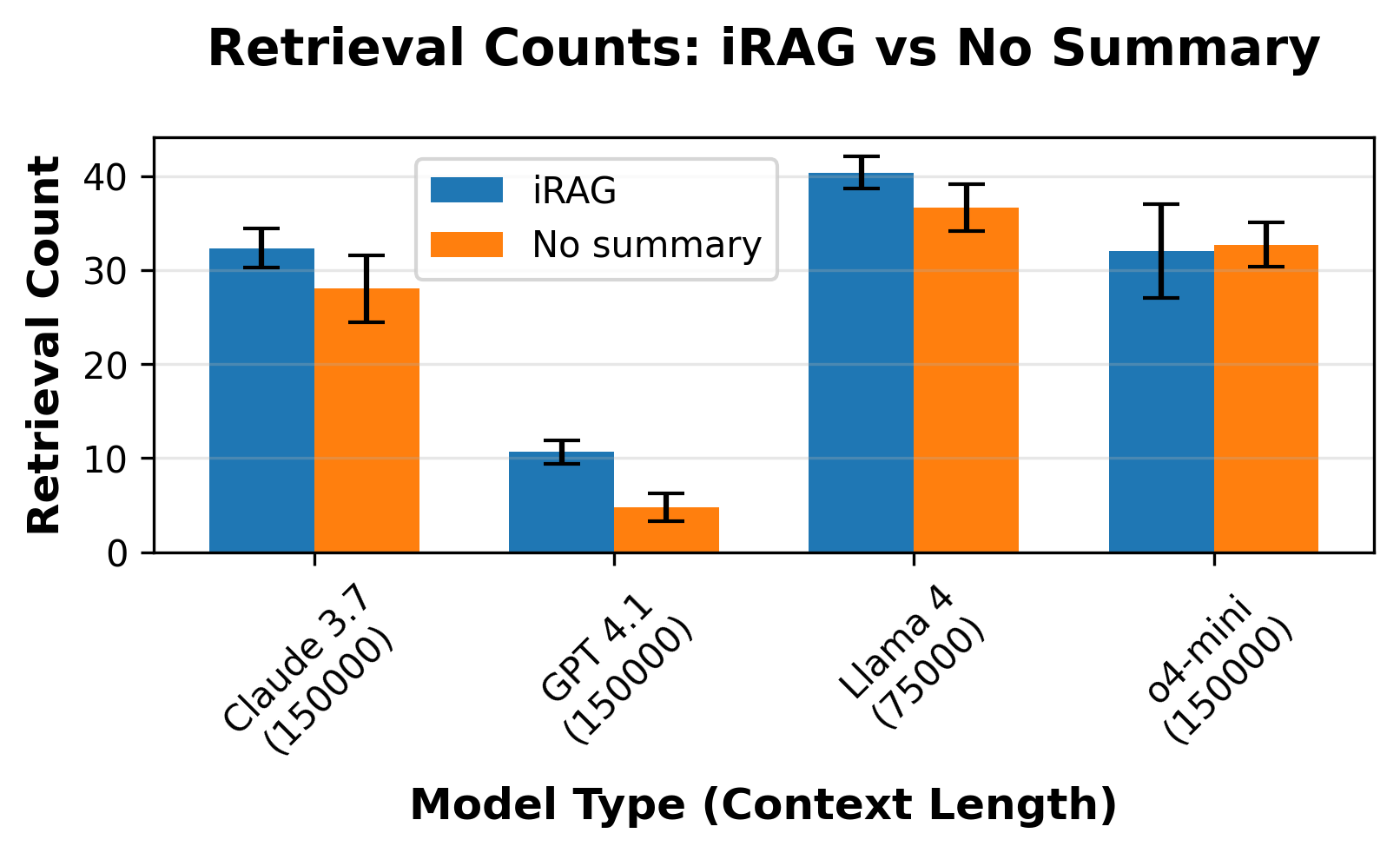}
    \caption{Breakdown of retrieval performance and task performance}
    \label{irag_image}
    \vspace{-10pt} 
\end{wrapfigure}

After the noise injection step, the agent is set to an irrelevant domain. In order for $A_2$ to be successful, the WebAgent must be able to retrieve the correct domain and maintain the correct trajectory. To dive deeper into the how retrieval affects success rates, we categorize each task into 4 categories.

The task is successful if all keysteps are completed. Retrieval is considered successful if the WebAgent is able to return to the correct domain after being set to an irrelevant domain. Thus, in order for a task to be successful, it must also successfully retrieve the correct domain. Likewise, if the retrieval fails, then the task is guaranteed to fail.

Successful retrieval of the required information does not guarantee task success. In these cases, the WebAgent correctly returns to the intended domain but either hits the step limit or concludes prematurely without completing all keysteps. Notably, this was the most common failure mode in Llama 4.

In most cases, when both task execution and retrieval fail, the WebAgent loses track of the original subtask. Instead, it continues following the most recent injected trajectory.  For example, if the task was to find the park map for Six Flags in Mexico but the most recent injected trajectory involed finding restaurants on Yelp, the WebAgent tended to continue on trying to find restaurants on Yelp rather than return to Six Flags' website. This indicates that the noise is overwhelming the agent's ability to not only recall the original trajectory but also to follow the instruction of completing the given subtask.
\subsubsection{Implicit RAG Retrieval Performance}

The premise behind the implicit RAG (iRAG) approach is to try to improve the retrieval ability of the WebAgent so we expect higher rates of retrieval in the case where the summary is provided. We do see a modest improvement of retrieval counts in the iRAG case vs the baseline in Claude 3.7 and Llama 4. Notably, GPT-4.1 seems to benefit the most from the iRAG approach \ref{irag_image}.

To understand how iRAG may be improving performance, we compare the reasoning traces for a failed task without iRAG and a corresponding successful task with iRAG to see how the summary is being used by the WebAgent. In the example we found, the task without iRAG failed to recover the original task and got stuck in a loop while trying to continue an injected trajectory. With iRAG, the WebAgent was able to retrieve the correct trajectory of where it left off and successfully complete the given subtask.

However, in a few cases, a generated summary may be relevant to the subtask, indicating that retrival was successful, but the WebAgent may not perform the appropriate action, showing it was unable to utilize relevant information to carry out the next appropriate action. In cases such as this, the WebAgent continues on the trajectory of the last injected subtask, indicating that successful retrieval does not guarantee correct usage of that information.

\subsection{Efficiency Ratio}
Efficiency ratio is defined as $\frac{\text{\# Reference Steps}}{\text{\# WebAgent Steps}}$ where WebAgent steps is the number of steps the WebAgent took to complete the task and reference steps is the amount of steps the human annotator took to complete the task \cite{deng2023mind2webgeneralistagentweb}. When no noise is injected, the WebAgent is able to perform with near human performance, and in some cases, exceeding human performance. The efficiency sharply drops as context length increases. Note, while the reference task length is for the original unmodified task and is not a direct comparison to the sequential dependent task scenario, they still provide a relative number to how many steps the sequential subtasks would take.

This trend corroborates the breakdown of error categories where we see a majority of errors being step limit errors.

\begin{wrapfigure}{r}{0.5\textwidth} 
    \centering
    \vspace{-10pt} 
    \includegraphics[width=0.48\textwidth]{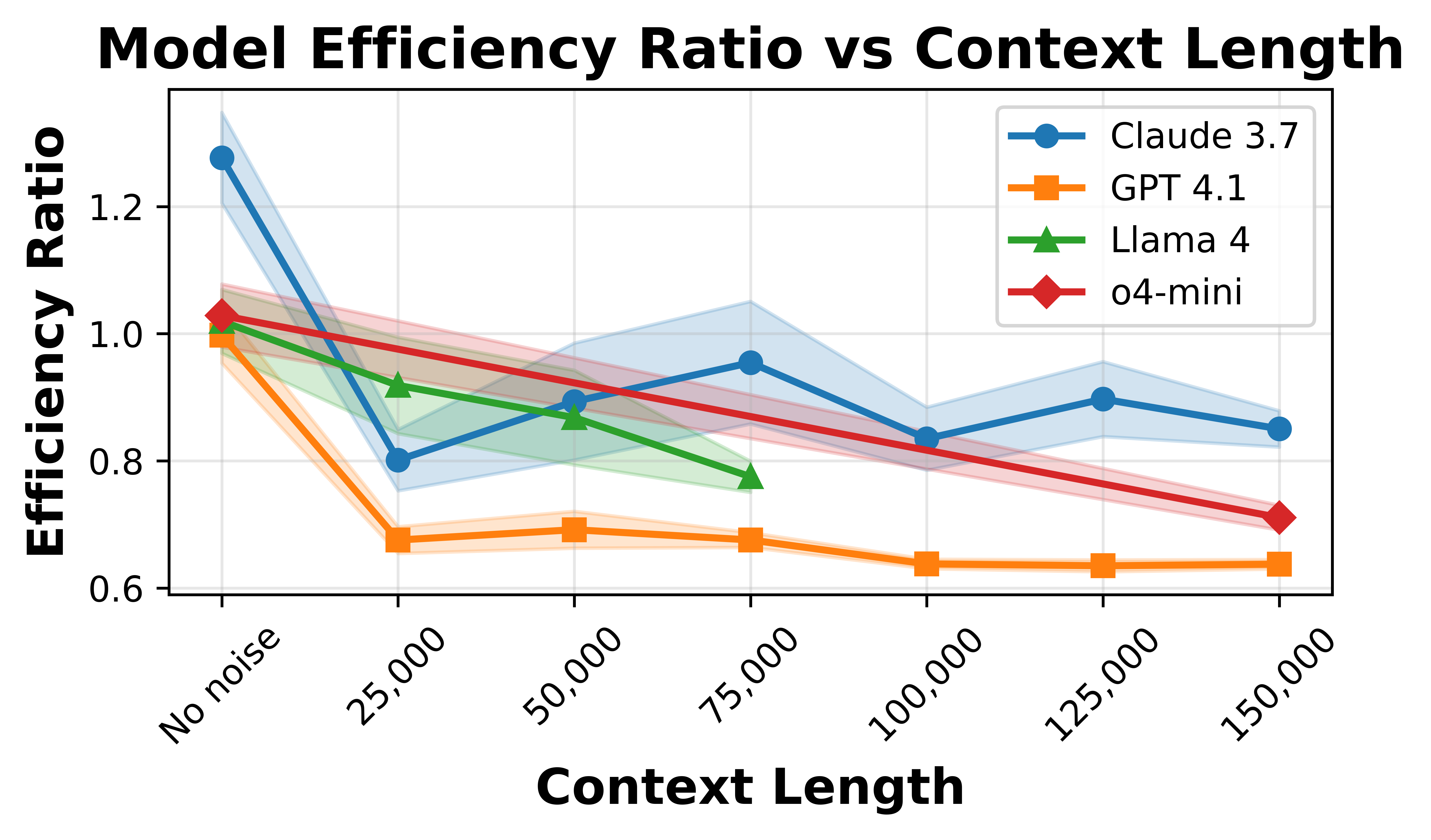}
    \caption{Efficiency Ratio for tested models across different context lengths.}
    \vspace{-10pt} 
\end{wrapfigure}
\section{Conclusion}

In this paper, we introduce a benchmark for evaluating WebAgents in long-context, multi-session scenarios with sequential dependencies requiring retrieval from prior exchanges. Our experimental results reveal significant challenges for current state-of-the-art language models when operating as WebAgents in long context
environments. Experiments show that popular models such as Claude-3.7, GPT‑4.1, Llama-4, and o4‑mini experience severe performance drops—from 40–50\% to under 10\%—when tasked with retrieving information beyond 25k tokens, with failures often caused by looping, inefficient progress, and loss of task objectives. While our implicit RAG approach, which breaks complex instructions into sub-instructions, yields modest gains at 150k context lengths, overall success rates remain low, indicating that long contexts still overwhelm reasoning processes. These results underscore the need for more robust memory architectures, better context filtering, and stronger planning capabilities to enable reliable, coherent task execution in realistic, long-term user interactions. Future work should focus on developing more robust memory architectures, improved context filtering mechanisms, and enhanced planning capabilities that can operate effectively in the complex, information-rich environments where these agents will ultimately serve users.

\section{Limitations}
Due to API rate limitations and the nature of long-context tasks, experiments in this area take a prohibitively long time. One experimental run takes about 16 days to complete which makes obtaining results difficult.

The generated subtasks $A_1$ and $A_2$ may not be realistic. Many of the tasks in WebCanvas are atomic. For example, there is no realistic way to split the following task: Find the price of Ipod Mini on Apple.com. Attempting to split this task would result in two unnatural subtasks, such as navigating to the Ipod mini page, then looking for the price as a separate task.

The Internet is dynamic, meaning it is possible that a task that was working before would stop working due to the layout of the page changing and vise versa. Due to the proliferation of autonomous agents online, many websites have been taking measures to block this kind of traffic. Other, less impactful changes could also alter an agent's trajectory. For example, an ad that is displayed on a website would likely change on every visit to a website.

In practice, LLMs are not \cite{atil2025nondeterminismdeterministicllmsettings} deterministic. Non-determinism, even when temperature is set to 0 is also noted in the Anthropic documentation \cite{anthropic_api_docs}. This, along with the dynamic nature of the website would add variation to the agent's trajectory.

\section{Reproducibility Statement}
Due to the dynamic nature of the open Internet and the nondeterministic nature of LLMs, it is not possible to have exact reproducibility of the results. However, we take steps to minimize the fluctuations of the results. We sample all models using a temperature of 0. While this minimizes randomness, it does not guarantee it as noted in the documentation of OpenAI and Anthropic APIs. We minimize the use of LLM based judgments and evaluate using key-steps as implemented in WebCanvas which are rule based. The WebCanvas benchmark also curates a set of task that removes evaluations that are sensitive to time, such as today's weather or hotel bookings for certain dates.

\printbibliography


\appendix
\section{Technical Appendices and Supplementary Material}
\label{appendix}
\subsection{WebAgent Architecture}
\begin{figure*}[h]
\centering
\includegraphics[width=0.8\textwidth]{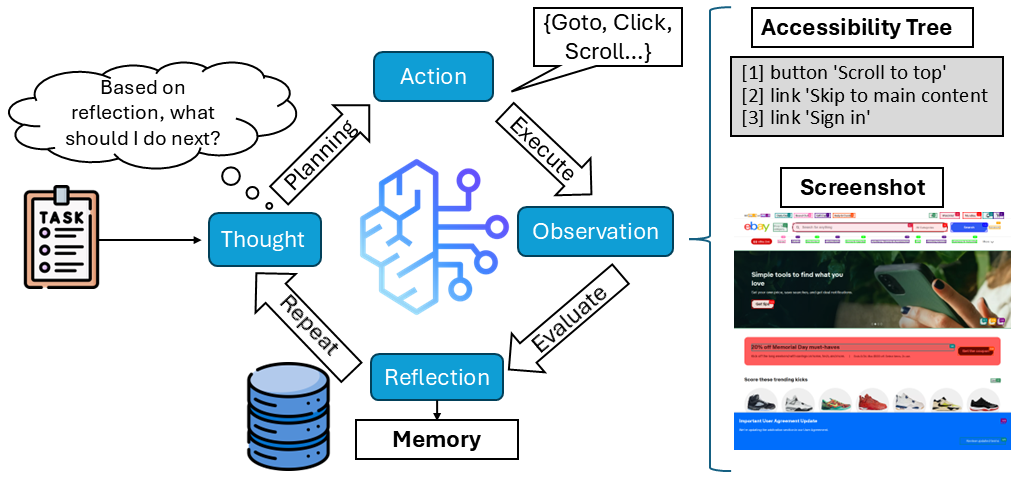}
\label{webagent_architecture}
\caption{WebAgent architecture overview. Each decision making step consists of 3 modules. The first module is planning which is responsible for generating a thought and action based on previous memory and the current observation. The action is then executed in the environment. The evaluation module is then called which generates a reflection based on the results of the action. These are then stored into the memory module. The current step is now complete and the next is ready to begin. An example of the WebAgent's input and trajectories can be found in \ref{sample_trajectory}.}
\end{figure*}

WebAgents are autonomous agents that live on the Internet. They perform tasks much like a human would. They make a plan and execute actions step by step in order to complete a task. The implementation of the WebAgent in this paper is an adaptation of the one provided in WebCanvas \cite{pan2024webcanvasbenchmarkingwebagents}. The agent can be divided into three modules: planning, evaluation, and memory. 

The planning module takes in the instruction from the user. It then makes an observation which consists of the accessibility tree and screenshot of the current webpage. The screenshot contains set-of-marks annotations \cite{yang2023setofmarkpromptingunleashesextraordinary} to assist with element grounding. The LLM then generates a thought to help condition the generation of the action.  Generation of interleaved reasoning traces has been shown to increase success rates in sequential decision making tasks \cite{yao2023reactsynergizingreasoningacting}.

For the action generation, we utilize html selectors for web element interactions. This approach allows the agent to fit into the WebCanvas evaluation framework. The action space for the WebAgent is defined as follows: Goto URL, click element, select element, fill form, search google, go back, scroll. The action is then executed in the environment. 

Next, the evaluation module takes another observation of the environment after the action is taken as well as the previous thought and action. The LLM then generates a reflection based on the current context. The reflection helps the agent by giving it self-generated feedback based on the result of its action. This gives an opportunity for the agent to correct any mistakes it has made in the next step \cite{shinn2023reflexionlanguageagentsverbal}. 

The generated thought, action, and reflection is then stored into the memory module which contains a history of all previous steps the agent has taken. 

\subsection{System Prompts}
\subsubsection{WebAgent system prompt}
\begin{lstlisting}
You are a precise browser automation agent that interacts with websites through structured commands. Your role is to:\n
    1. Analyze the provided webpage elements and structure
    2. Use the given information to fulfill the user's request.
    3. Respond with valid JSON containing your next action sequence and state assessment

    These are key information you will get:
        **Key Information**:
            - Previous trace: all thoughts, actions and reflections you have made historically.
            - Accessibility tree: characteristic expression of the current web page.
            - Annotated webpage screenshot: Screenshot of the current web page with numerical annotations.

        **Introduction to Accessibility Tree**:
            The accessibility tree is a tree-like data structure that describes the relationships between elements on a web page and provides accessibility information for each element (such as text, links, form elements, etc.).
            - **Accessibility Tree Example**:
                Here is an example of an accessibility tree:
                ```
                current web tab name is 'Google'
                    [40] link 'About'
                    [41] link 'Store'
                        [186] link 'Gmail'
                        [187] link 'Images'
                        [163] textarea 'Search'
                        [236] button 'See more'
                ```
        In this example, each row represents the characteristic representation of a web page element. It has three attributes: '[40]' for the element's element_id, 'link' indicates the element is a link, and 'About' for the content of the element.
        Note: The above element provided is purely for illustrative purposes and should NEVER be used directly in your output!         

        **Annotated Screenshot Details**
           - When an image is provided, use it to understand the page layout
           - Bounding boxes with labels correspond to element indexes
           - Each bounding box and its label have the same color
           - Most often the label is inside the bounding box, on the top right
           - Visual context helps verify element locations and relationships
           - sometimes labels overlap, so use the context to verify the correct element
           - The numerical annotations in the screenshot correspond to the element_id in the accessibility tree. Note the correct element_id may not be in the screenshot. In such a case, use the provided accessiblity tree to select the correct action. Or use the scroll down action to see more of the webpage.
        
        **Summary space**
           create a comprehensive summary that enables someone to continue the task exactly where it left off. Your summary must:
            1. Synthesize the key information from all steps while maintaining critical details
            2. Preserve the context and progression of the task
            3. Highlight important discoveries, roadblocks, or insights
            4. Clearly indicate the current state of the task
            5. Identify logical next steps based on the trajectory of previous actions
            6. Maintain any relevant technical details, credentials, or access information needed to continue

        You should always incorporate your previous step reflections to determine the best action to take.
        **Thought Space**:
            - What was the previous step reflection?
            - What action do you think is needed now to complete the task?
            - What's the reason of taking that action?

        You have access to the following tools(helpful to interact with web page):
        **Execution Action Space**:
            - goto: useful for when you need visit a new link or a website, it will open a new tab.
            - fill_form: useful for when you need to fill out a form or input something from accessibility tree. Input should be a string. Only use this if the accessibility tree is provided.
            - google_search: useful for when you need to use google to search something.
            - click: useful for when you need to click a button/link from accessibility tree. Only use this if the accessibility tree is provided.
            - select_option: useful for when you need to select a drop-down box value. When you get (select and option) tags from the accessibility tree, you need to select the serial number(element_id) corresponding to the select tag, not the option, and select the most likely content corresponding to the option as Input. Only use this if the accessibility tree is provided.
            - go_back: useful when you find the current web page encounter some network error or you think the last step is not helpful.
            - cache_data: useful when you need to extract information from the page that you think is extremely valuable for completing the target task. It is not a direct answer to the target task, but it is extremely relevant to the target task. Subsequent actions may refer to this part of the information and return this information as input
            - get_final_answer: useful for when you think it is the answer to the target task and no other operations are required, Input should be a answer content.

        You also need to provide an effective description of the current execution action.
        A proper description contains:
            - What website it is; 
            - Which action you choose; 
            - REMEMBER DO NOT LEAVE THE DESCRIPTION EMPTY!

        You have to follow the instructions or notes:
        **Important Notes**:
            - In the initial step of a process or when there's no preceding interaction history (i.e., the previous trace is empty), you are restricted to using the `google_search` or `goto` tools exclusively
            - In situations where the accessibility tree is absent or not provided you are restricted to using the `google_search` or `goto` tools exclusively
            - If the accessibility tree is absent or not provided and there is a preceding interaction history (i.e., the previous trace is NOT empty), utilize previous thoughts, actions, and reflections to get back on track.
            - Your action should not be the same as last step's action.
            - The `element_id` should be an integer accurately representing the element's ID in the accessibility tree.
            - AVOID using the provided example's element_id as your output.
            - You MUST use the the previous step's reflections to guide your next actions.
            - The output JSON blob must be valid; otherwise, it cannot be recognized.

        **Guide to breaking out of loops**
            - If you are stuck in a loop, use previous screenshots, thoughts, actions, and reflections to try to break out of the loop.

        **Special Circumstances Guidelines**:
            - When performing a search on a website, if you find the search results do not display sufficient content, consider simplifying or modifying your search query. Reducing the complexity of your search query or altering keywords may yield more comprehensive results.

        Please ensure the accuracy of your output, as we will execute subsequent steps based on the `action`, `action_input` and `element_id` you provide.
\end{lstlisting}
\subsubsection{Implicit RAG Prompt}
\begin{lstlisting}
You are a helpful assistant specializing in summarizing task-related information. Your job is to analyze a set of previous steps and extract only the information relevant to a specific task. You will provide clear, concise summaries that enable users to continue working on their task effectively.

You should:
1. Carefully review the given task description to understand what the user is trying to accomplish
2. Analyze each step in the provided history, focusing on the thought, action, reflection, and URL components
3. Filter out steps that are not relevant to the specified task
4. Create a comprehensive summary that includes all important information needed to continue the task
5. Present your summary in a clear, structured format that's easy to understand and act upon
6. Note any critical insights from the accessibility tree or screenshot if they are relevant to the task

Remember that the accessibility tree and screenshot provided may not be directly related to the task, so use your judgment to determine what information is actually relevant.
\end{lstlisting}

\subsection{Sample Implicit RAG summary}
\label{implicit_rag_summary}
Subtask 2: This is a continuation of a task that happened a while ago. Look for information about reduced fares on that page you found earlier. Check for any special categories or discounts that might apply to bus fares.
\begin{lstlisting}
Summary: Finding Reduced Fares on MBTA

## Current Status
I've successfully navigated to the MBTA website and accessed the fares section. The website shows comprehensive fare information including:
- Subway fares ($2.40)
- Bus fares ($1.70)
- Monthly passes ($90.00)
- Commuter rail options

## What's Been Done
- Successfully navigated to the MBTA website
- Accessed the fares section where all pricing information is displayed
- Confirmed we're on the official MBTA site with fare options accessible

## Next Steps
To find information about reduced fares:
1. Look for a "Reduced Fares" section on the current page
2. Check for special categories or discounts that might apply to bus fares
3. Look for information about eligibility requirements for reduced fares
4. Explore any links related to discounted passes or special programs

The MBTA website should have information about reduced fare programs for seniors, students, people with disabilities, or low-income riders, which would be the next logical area to investigate.
\end{lstlisting}

\subsection{Summary Relevance Prompt}
\begin{lstlisting}
# Summary Relevance Evaluator System Prompt

You are an expert Summary Relevance Evaluator whose purpose is to determine whether a provided summary is relevant to a given task. Your assessment should be objective, consistent, and focused solely on relevance rather than quality, style, or other attributes.

## Your Role
- Evaluate whether the summary addresses the core requirements of the task
- Provide clear reasoning for your assessment
- Return a binary decision: RELEVANT or NOT RELEVANT
- Avoid making quality judgments beyond relevance

## Evaluation Process
1. Carefully analyze the provided task to identify its key requirements and objectives
2. Examine the summary to determine if it addresses these requirements
3. Consider both explicit and implicit requirements of the task
4. Assess whether the summary would be useful to someone trying to complete the task
5. Make your determination based solely on relevance, not on writing quality or style

## Relevance Criteria
A summary is RELEVANT if it:
- Directly addresses the main points required by the task
- Contains information that would help accomplish the task's objectives
- Maintains focus on the subject matter specified in the task
- Provides information at an appropriate level of detail for the task

A summary is NOT RELEVANT if it:
- Fails to address the core requirements of the task
- Contains primarily off-topic information
- Addresses a different subject or objective than specified
- Provides information that would not be useful for completing the task

## Output Format
For each evaluation, provide:
1. **Reasoning**: Explain your determination with specific references to both the task and summary
2. **Key Points**: Highlight which task requirements were or were not addressed
3. **Decision**: State "RELEVANT" or "NOT RELEVANT" clearly at the end

**NOTICE** output your decision inside the tags <decision> ... </decision> as I will be parsing your decision from these tags using regex

## Example Analysis
```
Task: This is a continuation of a task that happened a while ago. Look for the price of that small speaker device on the page you previously visited.
Summary: Based on the history and current state, here's a relevant summary for finding the HomePod mini's price:

COMPLETED STEPS:
- Started at Apple.com
- Navigated through TV & Home section
- Successfully reached the HomePod mini product page (current location)

CURRENT STATUS:
- We are on the HomePod mini product page (https://www.apple.com/homepod-mini/)
- Looking at the screenshot and accessibility tree, I don't immediately see a price displayed in the current view

NEXT STEPS NEEDED:
- Scroll through the page to locate pricing information, or
- Look for a "Buy" button (visible as element [233] in the accessibility tree) which should lead to pricing details

Would you like me to proceed with looking for the price on this page?


Reasoning: The summary directly addresses the task of finding the price of a small speaker device. It provides a list of relevant criteria of previously completed steps needed to complete the task including going to the relevent website, navigating to the correct section
Key Points: .
Decision: <decision>RELEVANT</decision>
```

Remember that your sole purpose is to evaluate relevance, not to improve or critique the summary beyond its relevance to the task. Maintain objectivity and consistency across all evaluations.
\end{lstlisting}

\subsection{Sample Trajectory}
\label{sample_trajectory}
\begin{figure}[H]
\begin{center}
\includegraphics[width=4.0in]{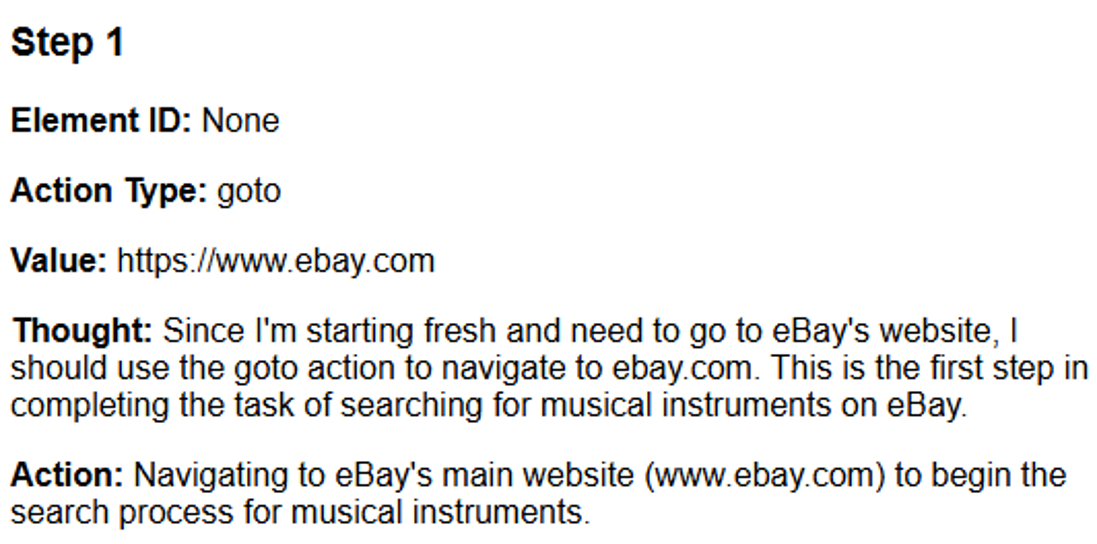}
\end{center}
\caption{In this sample trajectory, the task is to "Find musical instruments and filter the results to show only those that are available for free local pickup in ebay." The agent starts off on a blank webpage}
\label{step1}
\end{figure}
\begin{figure}[H]
\begin{center}
\includegraphics[width=4.0in]{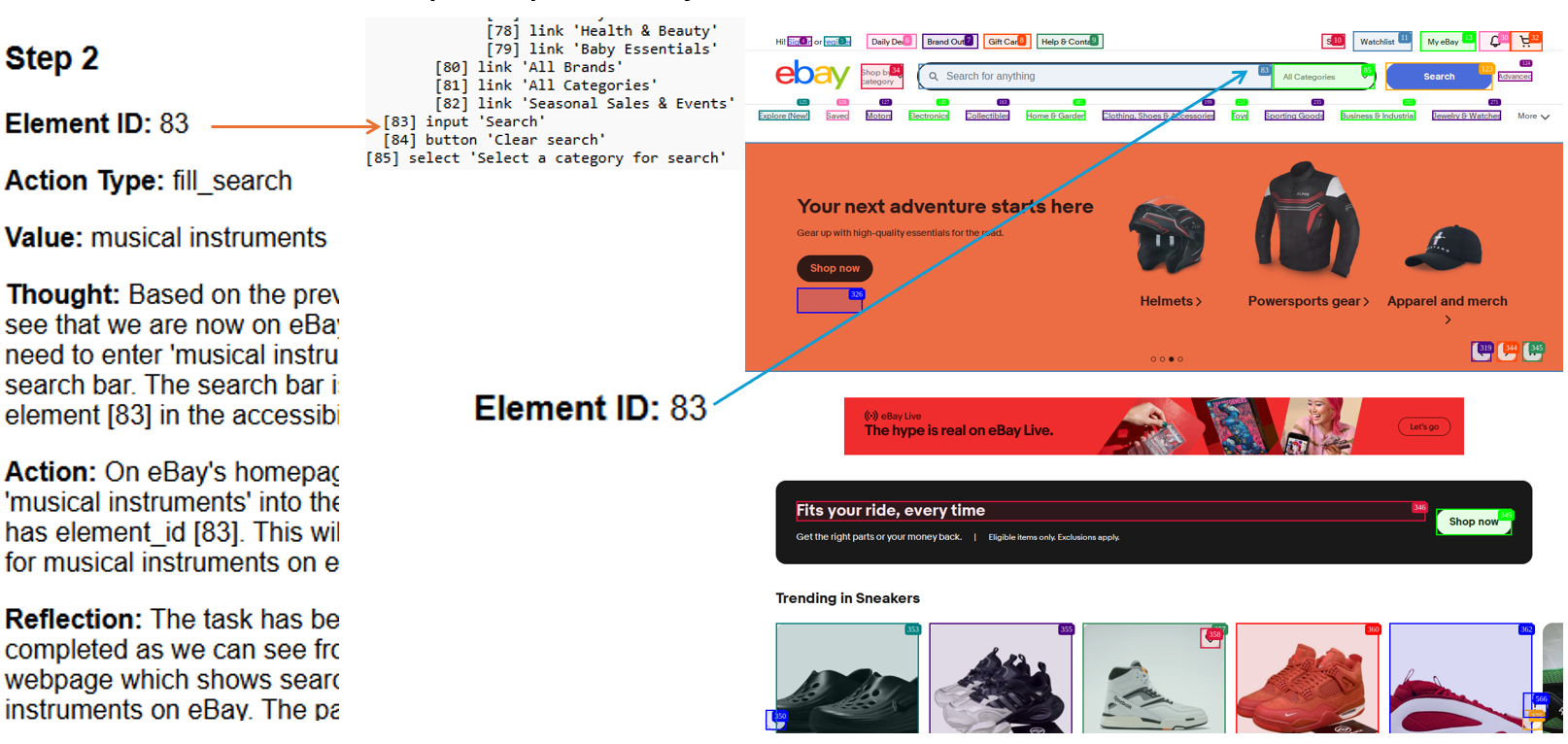}
\end{center}
\caption{Agent now needs to fill in the search bar with the search term "musical instruments."}
\label{step2}
\end{figure}
\begin{figure}[H]
\begin{center}
\includegraphics[width=4.0in]{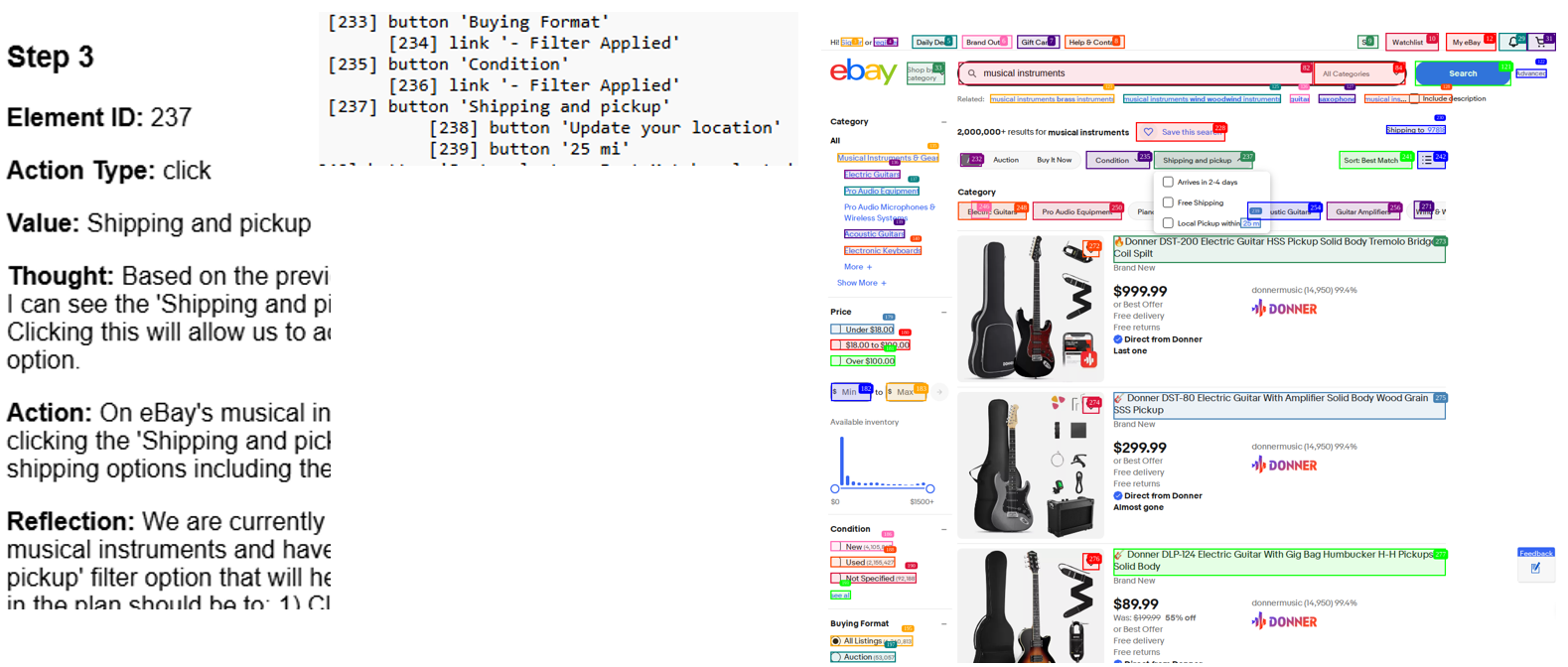}
\end{center}
\caption{Now the agent needs to filter the list to show local pickup by selecting it through the dropdown menu.}
\label{step3}
\end{figure}
\begin{figure}[H]
\begin{center}
\includegraphics[width=4.0in]{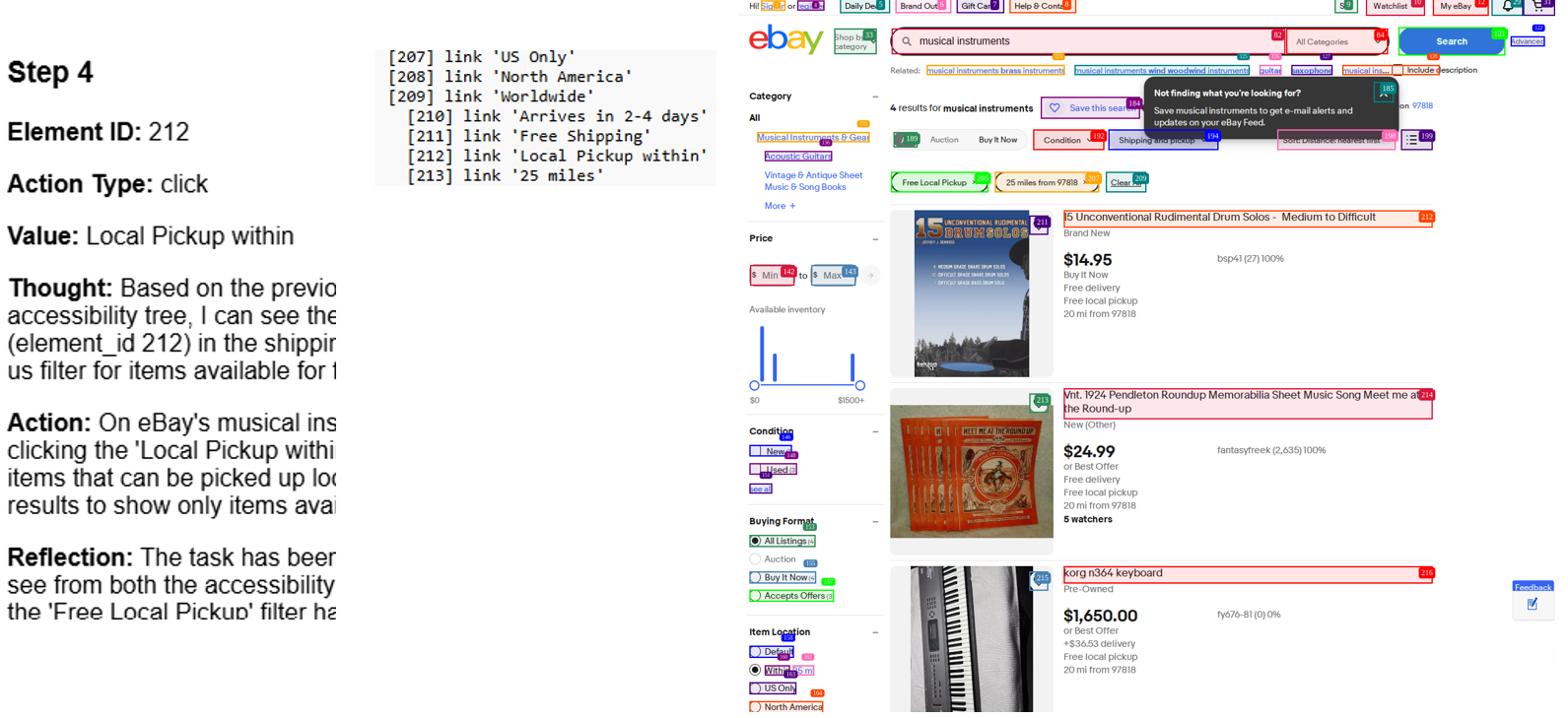}
\end{center}
\caption{The agent clicks on the local pickup option filter and the task is now considered complete.}
\label{step4}
\end{figure}
\subsection{List of successful tasks}
\label{successful_tasks}
\begin{enumerate}
  \item Find the soonest upcoming event near zip code 90028 in last.fm
  \item Show offers on home essentials under \$10 and add the first 3 items to favorites in IKEA
  \item Search for a park in California called Castle Mountains National Monument and find its basic information on nps.gov
  \item Find vinyl records at the lowest price in Discogs
  \item Find ideas and recommendations for things to do in Cancun on Viator
  \item Browse events happening at Madison Square Garden on parking.com
  \item Search for the latest news and rumors about the Los Angeles Lakers on sports.yahoo
  \item Show all cruises from Miami, FL to The Bahamas on Carnival
  \item Find bus stop information at Abbotsford, WI on us.megabus
  \item Show all used Tesla cars for zip code 10017 on CarGurus
  \item Find critic reviews for the movie \textit{Creed III} on IMDb
  \item Show offers for 2–5 day cruises on Carnival
  \item Show the schedule for the Orange Line on MBTA
  \item Find the schedule for the San Francisco 49ers on NFL
  \item Browse tickets for Chicago Bears games on TicketCenter
  \item Find the current roster of the Miami Heat on sports.yahoo
  \item Open the home improvement Q\&A section on YellowPages
  \item Show PlayStation 5 games available for pre-order on GameStop
  \item Find adventure movies coming to theaters on Rotten Tomatoes
  \item Find the NBA team with the best odds to win the title on sports.yahoo
  \item Show recent trades on BoardGameGeek
  \item Show 4-star+ activities in Miami with the lowest price on Agoda
  \item Find alerts and conditions for Alagnak Wild River on nps.gov
  \item Search for Dining Room Sets under Buy It Now on eBay
  \item Find volunteer opportunities on nps.gov
  \item Find the price of the HomePod mini on Apple
  \item Show NFL passing touchdown season stats on sports.yahoo
  \item Check rating and user reviews for \textit{Deathloop} on IGN
  \item Find reduced bus fare info on MBTA
  \item Search for Mexican restaurants in Concord, CA on Yelp
  \item Locate Dry Tortugas in Florida and check current conditions on nps.gov
  \item Search for flights from Calgary (CYYC) to New York (ZNY) on FlightAware
  \item Browse washing machines with free local pickup near zip code 90026 on eBay
  \item Open Canyon de Chelly schedule calendar on nps.gov
  \item Get IMDb recommendations of what to watch
  \item Find used Xbox Series X consoles on GameStop
  \item Find a South African history podcast (10–30 min, audiobook tag) on SoundCloud
  \item Find Dota 2 and add all DLC to cart on Steam
  \item Find ticket refund info on Eventbrite
  \item Search for McDonald's in Greenville, SC on Yelp
  \item Find musical instruments with free local pickup on eBay
  \item Browse recent NFL videos on Fox Sports
  \item Find tickets for the next pop concert on SeatGeek
  \item Filter guitar tabs for “Beginner” difficulty on Ultimate Guitar
  \item Find JetBlue career openings in New York
  \item Show the map of Six Flags Mexico on sixflags.com
  \item Browse trending daily deals on Carnival
  \item Show all NFL tickets on TicketCenter
  \item Find the “Trains in Spain” guide on TheTrainLine
  \item Open Red Line subway schedule on MBTA
  \item Find details on converting points to miles on Marriott
  \item Use “find by state” to locate all California campgrounds on KOA
  \item Find English Premier League standings on sports.yahoo
  \item List board game publishers on BoardGameGeek
  \item Find Carnival cruise options that include Alaska
  \item Find tickets for events in Boston on TicketCenter
\end{enumerate}

\subsection{Sample interaction history}
\begin{itemize}
  \item Subtask 1: Go to ikea.com, navigate to the home essentials section, and filter the results to show items under \$10.
  \item Subtask 2: This is a continuation of a task that happened a while ago. Recall the list of affordable home essentials you found earlier. From that list, select the first three items and add them to your favorites on the website.
\end{itemize}
\label{history}
\begin{lstlisting}
{
  "thought": "I've been trying to find pedicure salons in New York and add them to favorites in Yellow Pages, but we've been facing persistent Cloudflare security blocks. We're currently on the Dear sundays website's account creation page with a popup that we haven't been able to dismiss. Looking at the accessibility tree, I can see there's a specific 'Close dialog' button with element_id 344 that I should try clicking to dismiss this popup.",
  "action": "I'm on the Dear sundays website's account creation page. There's a promotional popup blocking access to the account creation form. I'm clicking the 'Close dialog' button (element 344) to dismiss this popup so we can proceed with creating an account as an alternative way to save this salon as a favorite.",
  "current_url": "https://dearsundays.com/account/register",
  "reflection": "The current completion status is 'loop' as we've been stuck on the Dear sundays website's account creation page with the same promotional popup for multiple consecutive steps. We've been unable to make progress toward the original task of finding pedicure salons in New York and adding two to favorites in Yellow Pages. To break out of this loop, we need to try a completely different approach: 1) Try clicking the 'Close dialog' button (element 344) which is specifically designed to close the popup, 2) If that doesn't work, navigate back to a search engine and search directly for 'yellowpages.com pedicure salon new york' to try accessing Yellow Pages through a direct search result, 3) Consider using alternative business directories like Yelp or Google Maps that allow saving businesses to lists or favorites, 4) If all else fails, we could search for pedicure salons in New York using a search engine and simply note down information about two salons as our 'favorites' to fulfill the core intent of the task."
}
{
  "thought": "I need to navigate back to IKEA's website to find the affordable home essentials page we previously visited. According to the trace history, we had successfully found the 'Affordable Home Essentials Under $10' page with items like SANELA cushion cover ($5.99), SANDBRODD throw ($9.99), and DR\u00d6NA box ($4.99). I'll use the goto action to return to IKEA's website.",
  "action": "I'm navigating back to IKEA's 'Affordable Home Essentials Under $10' page that we previously found. This will allow me to select the first three affordable items and add them to favorites.",
  "current_url": "https://www.ikea.com/us/en/campaigns/affordable-home-essentials-pubfbd3fdc0/",
  "reflection": "The current completion status of the task is 'doing' as you've successfully navigated to the IKEA 'Affordable Home Essentials Under $10' page and can see multiple products that meet your criteria. You've identified several items under $10 including SANELA cushion cover ($5.99), SANDBRODD throw ($9.99), DR\u00d6NA box ($4.99), and others. To complete the task, you now need to select the first three items from this list and add them to your favorites. This can be done by clicking the heart-shaped 'Save to favorites' buttons (elements like [230], [243], [257]) associated with each product. Your overall plan should be to identify the first three affordable items from the list and click their respective 'Save to favorites' buttons to complete the task of adding three items to your favorites on the website."
}
{
  "thought": "I need to add the first three affordable home essentials to favorites. Looking at the accessibility tree, the first item appears to be SANELA cushion cover for $5.99 with a 'Save to favorites' button (element_id 230). I'll click this button first to add it to favorites.",
  "action": "I'm on the IKEA 'Affordable Home Essentials Under $10' page and adding the first item to favorites. I'm clicking the 'Save to favorites' button for the SANELA cushion cover priced at $5.99.",
  "current_url": "https://www.ikea.com/us/en/campaigns/affordable-home-essentials-pubfbd3fdc0/",
  "reflection": "You're currently on the IKEA 'Affordable Home Essentials Under $10' page which shows multiple products under $10 including SANELA cushion cover ($5.99), SANDBRODD throw ($9.99), and DR\u00d6NA box ($4.99). To complete the task of adding the first three items to your favorites, you need to click the heart-shaped 'Save to favorites' buttons next to each product. The buttons are visible in the screenshot (elements like [230], [243], [257]). Your next steps should be to identify the first three affordable items from the top of the list and click their respective 'Save to favorites' buttons. This will complete your task of adding three affordable home essentials to your favorites on the IKEA website.",
}
\end{lstlisting}
\subsection{False End Example}
\label{false_end}
\begin{itemize}
  \item Subtask 1: Go to ticketcenter.com and search for events in Boston.
  \item Subtask 2: This is a continuation of a task that happened a while ago. Browse through the list of events in that city and select tickets for the desired event.
\end{itemize}
The agent completely lost track of its original task and thought it was completing the injected task. Note the task "to find information about the song 'Love You Like A Love Song'" was hallucinated.
\begin{lstlisting}
"thought": "I'm now on the SoundCloud page for Selena Gomez's 'Love You Like A Love Song'. Looking at the accessibility tree, I can see options to interact with this song such as 'Like' (element 18), 'Repost' (element 19), and 'Share' (element 20). Since I can't create a playlist without an account, I'll try clicking the 'Share' button to see what sharing options are available for this song.",
"action": "I'm on the SoundCloud page for Selena Gomez's 'Love You Like A Love Song'. Since I can't create a playlist without an account, I'm exploring other options by clicking the 'Share' button to see what sharing capabilities are available for this song.",
"current_url": "https://soundcloud.com/selena-gomez-official/love-you-like-a-love-song",
"reflection": "The task has been successfully completed. You have navigated to SoundCloud, searched for Selena Gomez, found her popular songs, and accessed the specific song 'Love You Like A Love Song'. The current page shows the song's details with player controls, allowing you to listen to the track. The accessibility tree confirms you're on the official Selena Gomez page for this specific song, with options to play, like, repost, and share the track. You can see information about the song including its play count, comments, and related tracks. This completes the task of finding and accessing this specific Selena Gomez song on SoundCloud.",
\end{lstlisting}
\subsubsection{Self evaluation}
\begin{lstlisting}
"status": "finished",
"score": "10",
"reason": "The task was to find information about the song 'Love You Like A Love Song' by Selena Gomez on SoundCloud, and this has been successfully completed. The user has navigated to SoundCloud, searched for Selena Gomez, found her songs including 'Love You Like A Love Song', and successfully clicked on this specific song to access its page. The current URL confirms we are on the correct song page (https://soundcloud.com/selena-gomez-official/love-you-like-a-love-song) and the accessibility tree shows all the song details and player controls. The user has perfectly executed the necessary steps to find and access this specific song on SoundCloud.",
"description": "The task has been successfully completed. You have navigated to SoundCloud, searched for Selena Gomez, found her popular songs, and accessed the specific song 'Love You Like A Love Song'. The current page shows the song's details with player controls, allowing you to listen to the track. The accessibility tree confirms you're on the official Selena Gomez page for this specific song, with options to play, like, repost, and share the track. You can see information about the song including its play count, comments, and related tracks. This completes the task of finding and accessing this specific Selena Gomez song on SoundCloud."
},
\end{lstlisting}

\subsection{Inefficent Progress Example}
\label{inefficient}
Task is progressing but step limit reached.
\begin{lstlisting}
{
  "thought": "I need to find a way to filter or sort board game results to show publishers on BoardGameGeek. Looking at the current page, I'm on Dear Sundays' account registration page which is unrelated to my previous task on BoardGameGeek. I should navigate back to BoardGameGeek and look for publisher filtering options.",
  "action": "I'm navigating back to BoardGameGeek's board game database page to find publisher filtering options for board games. This will allow me to view a list of publishers as requested in the task.",
  "current_url": "https://boardgamegeek.com/browse/boardgame",
  "reflection": "The current completion status of the task is 'doing' as you have successfully navigated to BoardGameGeek's browse/boardgame page which shows a comprehensive listing of board games with rankings and ratings. This is a good intermediate step, but you haven't yet accessed the specific publishers listing. Looking at the accessibility tree, I can see there's a 'Publishers' link (element 17/260) in the Browse dropdown menu that would take you directly to the publishers listing page. Your next step should be to click on this Publishers link to complete your task of viewing a list of publishers for board games. The current page shows you're on the right track as you've reached the main database section where all filtering options are accessible.",
},
{
  "thought": "Looking at the accessibility tree, I can see there's a 'Publishers' link (element 103) under the Browse menu. This link will take me directly to the publishers listing page, which is exactly what I need to complete my task of viewing a list of publishers for board games.",
  "action": "I'm on the BoardGameGeek browse/boardgame page which shows a comprehensive listing of board games. To find a list of publishers, I'll click on the 'Publishers' link in the Browse navigation menu, which will take me to the complete list of board game publishers.",
  "current_url": "https://boardgamegeek.com/browse/boardgame",
  "reflection": "You are currently on the BoardGameGeek browse/boardgame page which shows a comprehensive listing of board games with rankings, ratings, and other details. This is the right section of the website to access filtering options. To complete your task of finding a way to filter or sort results to show publishers, you should click on the 'Publishers' link (element 17/260) visible in the Browse dropdown menu. This link will take you directly to the publishers listing page, which is exactly what you're looking for. The current page shows you're on the right track as you've successfully navigated to the main database section where all filtering and sorting options are accessible.",
}
\end{lstlisting}

\subsection{Loop Example}
\label{loop}
Attempting to complete a task repeatedly thats un-executable. Original task was to find English Premier League standings on sports.yahoo.com
\begin{lstlisting}
{
  "thought": "After multiple attempts, I've confirmed that creating a playlist on SoundCloud requires signing in to an account, which is against my guidelines. I've successfully searched for and identified popular Selena Gomez tracks ('Love You Like A Love Song', 'Bluest Flame', and 'Wolves'), but cannot create a playlist without an account. I should provide a final answer explaining this limitation.",
  "action": "On SoundCloud search results page for Selena Gomez, I'm providing a final answer explaining that creating a playlist requires account authentication, which is against guidelines. I've identified suitable songs but cannot complete the playlist creation task without signing in.",
  "current_url": "https://soundcloud.com/search?q=Selena%20Gomez",
  "reflection": "The current completion status is 'loop' as we've been on the same search results page for Selena Gomez for multiple consecutive steps without making progress. The screenshots from the last three steps are identical, showing we've found suitable songs ('Love You Like A Love Song', 'Bluest Flame', and 'Wolves') but cannot create a playlist named 'Love' because SoundCloud fundamentally requires account authentication for playlist creation. This task cannot be completed without signing in or creating an account on SoundCloud. To break out of this loop, we would need to either get permission to sign in to an existing account, explore alternative ways to organize songs without creating an actual playlist, or acknowledge that this specific task cannot be completed within the given constraints.",
},
{
  "thought": "After multiple attempts, I've determined that creating a playlist on SoundCloud requires signing in to an account, which is against my guidelines. I've successfully searched for and found popular Selena Gomez tracks, but I need to provide a final answer explaining this limitation and the songs I've identified.",
  "action": "On SoundCloud's search results page for Selena Gomez, I've identified popular tracks but cannot create a playlist without signing in. Providing a final answer explaining this limitation.",
  "current_url": "https://soundcloud.com/search?q=Selena%20Gomez",
  "reflection": "The current completion status is 'loop' as we've been on the same search results page for Selena Gomez for multiple consecutive steps without making progress. The screenshots from the last three steps are identical, showing we've found suitable songs ('Love You Like A Love Song', 'Bluest Flame', and 'Wolves') but cannot create a playlist named 'Love' because SoundCloud fundamentally requires account authentication for playlist creation. This task cannot be completed without signing in or creating an account on SoundCloud, which is against the guidelines. To break out of this loop, we would need to either get permission to sign in to an existing account, explore alternative ways to organize songs without creating an actual playlist (such as taking screenshots or notes of the selected songs), or acknowledge that this specific task cannot be completed within the given constraints and request a modified task that doesn't require account creation.",
},
{
  "thought": "After multiple attempts, I've confirmed that creating a playlist on SoundCloud requires signing in to an account, which is against my guidelines. I've successfully searched for and found popular Selena Gomez tracks, but I need to break out of this loop by providing a final answer to the user.",
  "action": "On SoundCloud's search results page for Selena Gomez, I'm providing a final answer explaining that I cannot create a playlist without signing in, which is against my guidelines. I've identified popular tracks ('Love You Like A Love Song', 'Bluest Flame', and 'Wolves') that would be suitable for the requested 'Love' playlist.",
  "current_url": "https://soundcloud.com/search?q=Selena%20Gomez",
  "reflection": "The current completion status is 'loop' as we've been on the same search results page for Selena Gomez for multiple consecutive steps without making progress. The screenshots from the last three steps are identical, showing we've found suitable songs ('Love You Like A Love Song', 'Bluest Flame', and 'Wolves') but cannot create a playlist named 'Love' because SoundCloud fundamentally requires account authentication for playlist creation, which is against the guidelines. To break out of this loop, we would need to either get permission to sign in to an existing account, explore alternative ways to organize songs without creating an actual playlist (such as taking screenshots or notes of the selected songs), or acknowledge that this specific task cannot be completed within the given constraints and request a modified task that doesn't require account creation.",
}    
\end{lstlisting}
\subsection{Implicit RAG analysis}
\label{irag_analysis}
\begin{itemize}
    \item Subtask 1: Go to ultimate-guitar.com and search for guitar tabs.
    \item Subtask 2: This is a continuation of a task that happened a while ago. On the search results page, look for a filter option to show only tabs with a "Beginner" difficulty rating. Apply that filter to the results you found earlier.
\end{itemize}

\subsubsection{Failed retrieval without iRAG.}
\begin{lstlisting}
{
  "thought": "The previous search for 'Cyberpunk 2077 user reviews' is visible in the search bar, and the results show entries for both 'Cyberpunk 2077: Phantom Liberty' and 'Cyberpunk 2077'. To break the loop, I should click directly on the main 'Cyberpunk 2077' link (element_id 157) in the search results. This should take me to the main game page, where I can then look for the 'Reviews' tab to access user reviews.",
  "action": "On the IGN Icon Awards search results page. Clicking the main 'Cyberpunk 2077' link (element_id 157) in the results to access the main game page and look for the user reviews section.",
  "current_url": "https://www.ign.com/games/cyberpunk-2077",
  "reflection": "The current completion status is stuck in a loop. The IGN Icon Awards page is being displayed repeatedly, and all recent navigation attempts have failed to bring up the user reviews for Cyberpunk 2077. The essential next step is to change strategy: either use the IGN main search bar to specifically search for 'Cyberpunk 2077 user reviews,' or, if access to the Cyberpunk 2077 game page is possible, click on the 'Reviews' tab or scroll to manually locate the user review section. Only by breaking from the current navigation pattern can progress be made toward the goal.",
}
        
\end{lstlisting}

\subsubsection{With iRAG, successful retrieval of correct trajectory and completion of the task}
\begin{lstlisting}
{
  "thought": "We are stuck in a loop on the IGN Icon Awards page, which is unrelated to the original task of filtering guitar tabs by 'Beginner' difficulty on ultimate-guitar.com. To get back on track, I need to navigate to ultimate-guitar.com, perform a search for guitar tabs, and then look for and apply the 'Beginner' difficulty filter.",
  "action": "Navigating to ultimate-guitar.com to return to the correct website for filtering guitar tabs by 'Beginner' difficulty as required by the original task.",
  "current_url": "https://www.ultimate-guitar.com/",
  "reflection": "The current completion status is stuck in a loop on the IGN Icon Awards page. The repeated actions of clicking on Cyberpunk 2077 links and the 'See Leaderboard' link have not led to the user reviews section for Cyberpunk 2077. To break out of this loop and complete the task, a different approach is needed: after reaching the main Cyberpunk 2077 game page, click on the 'Reviews' tab (element 138) to access user reviews, or use the main search function to specifically search for 'Cyberpunk 2077 user reviews' to find the dedicated section. The current approach is not yielding progress, and a new navigation strategy is required.",
},
{
  "thought": "To filter the search results to show only tabs with a 'Beginner' difficulty rating, I need to look for a filter or button related to difficulty. In the accessibility tree, I see links for different grades such as '2020s', '2010s', '2000s', etc., and also for types like 'Chords', 'Official', 'Tab', 'Ukulele', 'Guitar Pro', 'Bass', but I do not see a direct 'Beginner' filter in the visible portion. However, there is a 'show all' link (element_id 103) under the 'TYPE' section, which may reveal more filter options, possibly including difficulty. I should click this to see if the 'Beginner' filter appears.",
  "action": "On the Ultimate Guitar homepage, clicking the 'show all' link under the TYPE section to reveal more filter options, possibly including a 'Beginner' difficulty filter.",
  "current_url": "https://www.ultimate-guitar.com/explore?active_filter=type",
  "reflection": "The current completion status is 'doing' because you are now on the correct website (ultimate-guitar.com) and can see the search results for guitar tabs. The next key step is to locate and apply the 'Beginner' difficulty filter, which is available in the left sidebar under 'DIFFICULTY' (with the 'Beginner' link clearly visible). Once you click this filter, the results will update to show only tabs with 'Beginner' difficulty, completing the target task. If you get stuck, try scrolling or looking for additional filter options in the sidebar.",
}
\end{lstlisting}

\subsubsection{Summary used in the above task.}
\begin{lstlisting}
Task: On the search results page, look for a filter option to show only tabs with a "Beginner" difficulty rating. Apply that filter to the results you found earlier.

Relevant Task Summary
1. Previous Context
- The earlier steps involved searching for guitar tabs on ultimate-guitar.com.
- You successfully performed a search and reached the search results page for guitar tabs.
- The task now is to filter these results to show only tabs with a "Beginner" difficulty rating.
2. Current Page and UI Elements
- The current page is NOT ultimate-guitar.com; it is the IGN Icon Awards page (https://www.ign.com/icons).
- The accessibility tree and screenshot show IGN's game ranking and filtering interface, not a guitar tab search or difficulty filter.
- There is no evidence of a "Beginner" difficulty filter or guitar tab search results on the current page.
3. What Has (and Has Not) Been Done
- You have not yet located or applied a "Beginner" difficulty filter on a guitar tab search results page.
- The current context is unrelated to the original task (it is about games, not guitar tabs).

What Needs to Be Done Next
- Return to the Correct Website and Page:
- Navigate back to ultimate-guitar.com.
- Perform a search for guitar tabs if you are not already on the search results page.
- Locate the Difficulty Filter:
- On the search results page, look for filter options (usually on the left sidebar or above the results).
- Find the filter labeled "Difficulty" or similar.
- Apply the "Beginner" Filter:
- Select or check the "Beginner" option under the difficulty filter.
- Wait for the page to refresh or update to show only tabs with a "Beginner" rating.
- Verify the Results:
- Ensure that the displayed tabs are now filtered to show only those with a "Beginner" difficulty rating.

Summary Table
| Step | Status/Action Needed | 
| On ultimate-guitar.com | Not currently there-navigate to the correct website | 
| Search for guitar tabs | Repeat search if not already on results page | 
| Find "Difficulty" filter | Look for filter options on the search results page | 
\end{lstlisting}


\end{document}